\pdfoutput=1

\documentclass[11pt]{article}

\usepackage{acl}

\usepackage{times}
\usepackage{latexsym}

\usepackage[T1]{fontenc}

\usepackage[utf8]{inputenc}

\usepackage{microtype}

\usepackage{graphicx}
\usepackage{caption}
 \usepackage{booktabs}
 \usepackage{multirow}
\usepackage{subcaption}
\usepackage{tablefootnote}
\usepackage{float}
\usepackage{comment}
\usepackage{amsmath}
\usepackage{natbib}

\usepackage{xcolor}

\title{One Configuration to Rule Them All? \\Towards Hyperparameter Transfer in Topic Models\\ using Multi-Objective Bayesian Optimization}

\author{Silvia Terragni$^{\diamondsuit\clubsuit}$, Ismail Harrando$^\spadesuit$, Pasquale Lisena$^{\spadesuit}$, Raphael Troncy$^{\spadesuit}$, Elisabetta Fersini$^{\clubsuit}$ \\$^{\diamondsuit}$Telepathy Labs, Zurich, Switzerland, \\
$^{\clubsuit}$University of Milano-Bicocca, Milan, Italy, \\
$^\spadesuit$EURECOM, Sophia Antipolis, France\\
$^{\diamondsuit}$ \texttt{silvia.terragni@telepathy.ai}, \\$^\spadesuit$\texttt{\{harrando, lisena, troncy\}@eurecom.fr}, \\ $^{\clubsuit}$\texttt{elisabetta.fersini@unimib.it} }

\makeatletter
\newcommand\footnoteref[1]{\protected@xdef\@thefnmark{\ref{#1}}\@footnotemark}
\makeatother

\begin{document}
\maketitle
\begin{abstract}
Topic models are statistical methods that extract underlying topics from document collections. When performing topic modeling, a user usually desires topics that are coherent, diverse between each other, and that constitute good document representations for downstream tasks (e.g. document classification). 
In this paper, we conduct a multi-objective hyperparameter optimization of three well-known topic models. The obtained results reveal the conflicting nature of different objectives and that the training corpus characteristics are crucial for the hyperparameter selection, suggesting that it is possible to transfer the optimal hyperparameter configurations between datasets.
\end{abstract}

\section{Introduction}
\label{sec:introduction}
Topic models~\cite{blei2012probabilistic} are statistical methods that extract the underlying themes (\textquotedblleft topics'') from a document corpus. Topics are often represented by sets of words that make sense together, e.g. the words \textquotedblleft cat, animal, dog, mouse'' may represent a topic about animals. Topic models' evaluations are usually limited to the comparison of models whose hyperparameters are fixed~\cite{benchmarkingneuraltm2021}. However, hyperparameters can affect models' performance and fixing them prevents the researchers from discovering the best topic model on a dataset. 

\citet{Terragni21_octis} proposed a first attempt to conduct a robust comparison between models, using single-objective hyperparameter optimization to automatically get the optimal hyperparameters. This has been later extended to multi-objective hyperparameter optimization~\cite{Terragni21_octis2}. Yet, they do not provide a thorough and extensive experimentation using either single or multi-objective optimization, which may reveal possibly competing objectives. 

In fact, topic models can be evaluated in several ways: from downstream tasks, such as document classification and information retrieval~\cite{Boyd-Graber17tmapplications}, to the analysis of the generated topics~\cite{Lau14,Chang09_readingtealeaves}. However, previous work~\cite{Nan19} noted that these objectives can be conflicting. For example, a topic model that produces good document representations may not be able to produce coherent and diverse topics simultaneously. 

To this purpose, we further explore this direction by using the Multi-Objective Bayesian Optimization (MOBO)~\cite{Paria19_multiobjective} approach to optimize the hyperparameters of three well-known topic models according to three criteria: topic coherence, topic diversity, and document classification. Exploring the space of the hyperparameters using MOBO leads to jointly discover the best models according to the three considered objectives and reveal the conflicting nature of different objectives. 

Moreover, we investigate whether the discovered best hyperparameter configurations are consistent across datasets having different characteristics. In other words, we verify if we can always transfer the optimal configuration of hyperparameters from a dataset to another one, or if the dataset features play a part in the hyperparameter transfer problem. 

\paragraph{Contributions.}
In this paper, we show that we can use MOBO to boost the performance of topic models according to multiple objectives. However, these objectives are often competing (Research Statement \textbf{RS \#1}), leading to a model that is excellent for a purpose but performs poorly for another objective. Thus, a user would need to find a balance between different and conflicting objectives. 

Our results also show that, in some cases, we can transfer the best hyperparameter configurations from a dataset to an unseen dataset (Research Statement \textbf{RS \#2}). This result suggests that we need to take into consideration the dataset features when performing the hyperparameter transfer.




\section{Related Work}
\label{sec:related-work}

\paragraph{Hyperparameter Optimization.} Researchers adopt different strategies to address the problem of hyperparameters selection for topic models. They usually select a priori values according to domain knowledge. Yet, in most cases, there is no knowledge of the topic distribution over the corpus. An alternative is to select the hyperparameter configuration using grid search techniques~\cite{Griffiths2004,selfTraining2017}. Grid search is simple and parallelizable but suffers from the curse of dimensionality, as the number of possible configurations grows exponentially with the number of hyperparameters and the range of possible values~\cite{Bergstra12}. 

Another option adopts fixed-point methods for estimating the hyperparameters of a model~\cite{wallach2008structured, Asuncion09}. The inference algorithm alternates between sampling the topics and inferring the hyperparameters. \newcite{Terragni21_octis} recently proposed OCTIS, a topic modeling framework which incorporates a black-box optimization approach, i.e.  Bayesian Optimization, to optimize the hyperparameters of topic models. Bayesian techniques can be superior to point estimates and grid search techniques~\cite{Snoek12practicalBO}. However, a single-objective strategy disregards the importance of the other objectives in topic models. Indeed, this work has been later extended to Multi-Objective optimization~\cite{Terragni21_octis2}, although in both works an extensive experimental campaign is missing. 

\paragraph{Multi-objective topic modeling.} Although the literature on topic modeling evaluation is rich and diverse~\cite{Chang09_readingtealeaves,Korencic18doccoherence,wallach2009evaluation}, only a few efforts have been made to tackle this task as a multi-objective optimization problem. \newcite{khalifa13} and \newcite{gonzalessantos21} investigate the use of multi-objective evolutionary algorithms for topic modeling to infer coherent topics by searching trade-offs with regards to three objectives: topic coverage, topic coherence, and perplexity. 
Other approaches \cite{vorontsov2015,ding18,veselova20} explicitly model additional objectives into the process of building the topic model, e.g. by adding extra terms to the optimization objective/loss function. 

Our approach differs from the previously presented methods because it is model-agnostic. 
 Moreover, we use hyperparameter optimization aiming to show that we need to compromise between different conflicting metrics and it is possible to zero-shot transfer the best hyperparameter configurations to unseen datasets.

\section{Multi-objective Optimization for Topic Models}
To investigate the role of multi-objective optimization in topic models, we optimize the topic models' hyperparameters by adopting a Bayesian Optimization (BO) strategy~\cite{mockus1978application, archettibayesian,Kandasamy20_dragonfly,Paria19_multiobjective}. BO is a Sequential Model-Based Optimization strategy that allows us to optimize the hyperparameters of a model requiring little prior knowledge. In a single-objective setting, BO uses the model's configurations evaluated so far to approximate the value of the objective performance metric with respect to the model's hyperparameters and then selects a new promising configuration to evaluate. BO is based on two key components: a probabilistic surrogate model aiming at approximating the performance metrics to optimize and an acquisition function (also named utility function or infill criterion) to select the next configuration.

Single-objective BO can be generalized to multiple objective functions~\cite{Paria19_multiobjective}, where the final aim is to recover the Pareto frontier of the objective functions, i.e. the set of Pareto optimal points. A point is Pareto optimal if it cannot be improved in any of the objectives without degrading some other objective. 
In this paper, we will use the multi-objective methodology presented in~\citeauthor{Paria19_multiobjective}. We refer the readers to \cite{Paria19_multiobjective} for additional details on multi-objective Bayesian optimization and to \cite{Terragni21_octis, Terragni21_octis2} to the use of hyperaparameter Bayesian Optimization in topic modeling. 

\paragraph{Objective functions.}
We consider three well-known objective functions that investigate different aspects of a topic model: the quality of the topics, the diversity of the topics, and the prediction capability of the model in a classification task. These three aspects are usually investigated in the topic modeling literature~\cite{Chang09_readingtealeaves,Dieng2020,Terragni2019}. All the considered functions must be maximized. 

\begin{itemize}
 \item \textbf{Topic Coherence} measures how much the top-$n$ words representing each topic are related. A well-known metric of topic coherence is NPMI~\cite{Lau14}, which is computed using Normalized Pointwise Mutual Information of each pair of words $(w_i, w_j)$ in the top-10 words of each topic. 

 \item \textbf{Topic diversity} measures how much the top\nobreakdash-$n$ words of a topic differ from another topic. We use a recent metric~\cite{bianchi_diversity} based on Ranked-Biased Overlap~\cite{webber2010similarity}. Topics with common words at different rankings are penalized less than topics sharing the same words at the highest ranks. We will refer to this metric as IRBO.

 \item The \textbf{predicting capability} refers to the capability of a topic model to produce a good document representation. The topic distribution of each document is used to train a classifier. Here, we use a polynomial SVM and we compute the Micro-F1 measure, i.e. the weighted average of the F1 measure for each class (each dataset has a number of classes, specified in Table~\ref{tab:dataset_properties}). We will refer to this metric as F1. 
\end{itemize}

\section{Hyperparameter Transfer}
The knowledge related to optimal hyperparameter configurations, which we acquire during the multi-objective optimization, can be transferred to an unseen dataset. We follow a simple and effective hyperparameter transfer approach, based on the work of \newcite{Feurer14metalearning}. 

Let $f_i(x)$ with $i=\{1, \ldots, N\}$ denote an objective function (here, $N=3$) and $\theta_i^1, \ldots, \theta_i^D$ denote the best hyperparameter configurations discovered by MOBO for the previously seen datasets $1, \ldots, D$ respectively. Each $\theta_i^d$ is composed of the~$t$ best hyperparameters configuration for the objective function $f_i$. In \citeauthor{Feurer14metalearning}, the authors define some dataset features, also called \textit{metafeatures}, and a similarity measure for each feature, thus allowing to initialize the surrogate model of the BO with best hyperparameter configurations of the dataset that is the most similar. 

In this paper, we follow an opposite direction to see if an optimal hyperparameter configuration is consistent across different datasets (RS \#2).  
We train a topic model on an unseen target dataset with the best hyperparameter configurations $\theta_i^d$ of previously seen dataset $d$ for the given objective function~$f_i$. If a configuration can be effectively transferred to every dataset (i.e. the best hyperparameter configuration transferred from a dataset to the target one and vice versa achieves performances that are close to the optimal ones), then it follows that the configurations are independent of the datasets' features. Otherwise, if some configurations do not transfer well on a target dataset, it implies that the hyperparameter configurations are dependent on the metafeatures.

\section{Experimental Setting}
\subsection{Models}
In our evaluation, we consider three distinct topic models, chosen to be the representatives of different categories of topic models~\cite{Stevens12manymodels,survey_neural_models}: classical probabilistic models, matrix factorization methods, and neural topic models. Due to their different formulations, all the considered models are controlled by different types of hyperparameters that we will detail in Section~\ref{subsec:hyperparams}.

\paragraph{LDA}  \cite[{Latent Dirichlet Allocation}]{Blei03} \footnote{\url{https://radimrehurek.com/gensim/models/ldamodel.html}} is a generative probabilistic model that describes a corpus of documents through a set of topics $K$, seen as distributions of words over a vocabulary $W$. A document is assumed composed of a mixture of different topics that follow a Dirichlet distribution, where a topic
drawn from this mixture is assigned to each word of the document. 

\paragraph{NMF} \cite[Non-Negative Matrix Factorization]{paatero1994positive}  \footnote{\url{https://scikit-learn.org/stable/modules/generated/sklearn.decomposition.NMF.html}} is a statistical method that reduces the dimensionality of the input corpus of $D$ documents, viewed as a matrix $M$ of shape $D\times|W|$, where $|W|$ represents the length of the vocabulary. It aims at decomposing $M$ as the product of two matrices $V$ and $H$, such that the dimension of $V$ is $|W|\times K$ and that of $H$ is $D\times K$. The decomposed matrices
must consist of only non-negative values.

\paragraph{CTM\footnote{This model has been designed for addressing a task of cross-lingual topic modeling, however it also outperforms several monolingual neural topic models.}} (Zero-shot Contextualized Topic Model)~\cite{bianchiCTM}\footnote{\url{https://github.com/MilaNLProc/contextualized-topic-models}} is a recent neural topic model, based on the Variational Autoencoder (VAE) architecture. The neural variational framework trains a neural inference network to directly map a contextualized document representation -- e.g. coming from BERT~\cite{Devlin19_bert}-- into a continuous $K$-dimensional latent representation. Then, a decoder network reconstructs the document Bag-of-Word representation by generating its words from the latent document representation. 

\subsection{Datasets}
We consider 6 different datasets: 20~NewsGroups (20NG)\footnote{\url{http://people.csail.mit.edu/jrennie/20Newsgroups/}}, AFP\footnote{\url{http://193.55.113.124/topic-model-api/dataset/afp_fr.tsv}}, BBC News (BBC)~\cite{greene06icml}, M10~\cite{lim2015bibliographic}, StackOverflow (SO)\footnote{\label{fn:sttm_dataset}\url{https://github.com/qiang2100/STTM}}, and SearchSnippets (SS)\footnoteref{fn:sttm_dataset}~\cite{phan08classification}. 
Moreover, the AFP dataset is in French, while the others are in English. We report the main statistics of the preprocessed datasets in Table~\ref{tab:dataset_properties} and the preprocessing details in the Appendix. The datasets are split in training (70\%), testing (15\%) and validation set (15\%). 

\begin{table}[h]
\centering
\small
\begin{tabular}{lrrrr}
\toprule
\textbf{Name} & \textbf{\# Docs}  & \textbf{\# Labels} & \begin{tabular}[c]{@{}c@{}} \textbf{\# Unique} \\ \textbf{words}\end{tabular} & \begin{tabular}[c]{@{}l@{}} \textbf{ Avg doc} \\ \textbf{length (std)}\end{tabular} \\ \midrule
\textsc{20NG} & 16,309 & 20 & 1612 & 48 (130) \\
\textsc{AFP} & 26,599 & 17 & 2686 & 156 (174) \\
\textsc{BBC} & 2,225 & 5 & 2949 & 120 (72) \\
\textsc{M10} & 8,355 & 10 & 1696 & 6 (2) \\
\textsc{SS} &12,295 & 8 & 4705 & 14 (5) \\
\textsc{SO} & 16,407 & 20 & 2257 & 5 (2)\\\bottomrule
\end{tabular}%
\caption{Characteristics of the studied datasets}
\label{tab:dataset_properties}
\end{table}
\subsection{Multi-objective Hyperparameter Optimization}
\label{sec:MOHO}
To simultaneously optimize topic quality (NPMI), topic diversity (IRBO) and classification (F1), we use the \textit{Dragonfly} library~\cite{Kandasamy20_dragonfly,Paria19_multiobjective} integrated into the topic modeling framework OCTIS~\cite{Terragni21_octis,Terragni21_octis2}. To obtain robust evaluations, we train each model 30 times and consider the median of the 30 evaluations as the evaluation of the function to be optimized. A number $n$ of initial configurations is randomly sampled via Latin Hypercube Sampling ($n$ equal to the number of hyperparameters to optimize plus~2 to provide enough configurations for the initial surrogate model to fit). The total number of BO iterations for each model is 125. We use Gaussian Process as the probabilistic surrogate model and the Upper Confidence Bound (UCB) as the acquisition function.

\subsection{Single-objective Hyperparameter Optimization}

Similarly to \ref{sec:MOHO}, we use the \textit{Dragonfly} to perform single-objective Bayesian Optimization to optimize each of the individual metrics, to compare the performance on individual metrics when switching to the MO setting. Again, we use the same number of iterations and hyperparameters search space.

\subsection{Hyperparameter setting}
\label{subsec:hyperparams}

\begin{table}[h]
\resizebox{0.95\linewidth}{!}{%
\begin{tabular}{@{}lp{3.2cm}p{3.9cm}@{}}
\toprule
\textbf{Model} & \textbf{Hyperparameter} & \textbf{Values/Range} \\ \midrule
All & Number of topics & {[}5, 150{]} \\
\midrule
\multirow{2}{*}{LDA} &  $\alpha$ prior & {[}$10^{-4}, 10${]} \\
 & $\beta$ prior & {[}$10^{-4}, 10${]} \\\midrule
\multirow{6}{*}{NMF} & Regularization factor & {[}0, 0.5{]} \\
 & L1-L2 ratio & {[}0,1{]} \\
 & \multirow{2}{*}{Initialization method} & random, nndsvd, nndsvda, nndsvdar \\
 & Regularization & V matrix, H matrix, both \\\midrule
\multirow{11}{*}{CTM} & \multirow{2}{*}{Activation function} & softplus, relu, sigmoid, leakyrelu, rrelu, elu, selu \\
 & Dropout & {[}0, 0.95{]} \\
 & Learn priors & true (1), false (0) \\
 & Learning rate & {[}$10^{-4}, 10^{-1}${]} \\
 & Momentum & {[}0, 0.9{]} \\
 & Number of layers & 1, 2, 3, 4, 5 \\
 & \multirow{2}{*}{Number of neurons} & 100, 200, 300, 400, 500, 600, 700, 800, 900, 1000 \\
 & \multirow{2}{*}{Optimizer} & adagrad, adam, sgd, adadelta, rmsprop \\ \bottomrule 
\end{tabular}%
}
\caption{Hyperparameters and ranges.}
\label{tab:ranges}
\end{table}
\begin{figure*}[h]
\centering
    \includegraphics[width=1\linewidth]{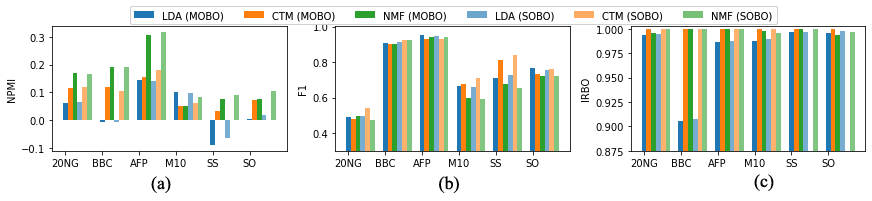}
     \caption{Best performance of the topic models for each evaluation metric on the considered datasets: (a) NPMI, (b) F1 (c) IRBO.}
     \label{fig:best_results}
\end{figure*}
We summarize the models' hyperparameters and their corresponding ranges in Table~\ref{tab:ranges}. For each model, we optimize the number of topics, ranging from 5 to 150 topics. Regarding LDA, we also optimize the hyperparameters $\alpha$ and $\beta$ priors that the sparsity of the topics in the documents and sparsity of the words in the topic distributions respectively. These hyperparameters are set to range between~$10^{-4}$ and $10$ on a logarithmic scale. 

The hyperparameters of NMF are mainly related to the regularization that can be applied to the factorized matrices. The \textit{regularization} hyperparameter controls if the regularization is applied only to the matrix $V$, or to the matrix $H$, or both of them. The \textit{regularization factor} is the constant that multiplies the regularization terms. It ranges between 0 and 0.5 (where 0 means no regularization). \textit{L1-L2} ratio controls the ratio between L1 and L2-regularization. It ranges between 0 and~1, where 0 corresponds to L2 only, 1~corresponds to L1 only, otherwise it is a combination of the two types. We also optimize the \textit{initialization method} for the matrices~$W$ and $H$.

Since CTM is a neural topic model, its hyperparameters are mainly related to the network architecture. We optimize the \textit{number of layers} (ranging from~1 to 5), and the \textit{number of neurons} (ranging from 100 to 1000, with a step of 100). For simplicity, each layer has the same number of neurons. We also consider different variants of \textit{activation functions} and \textit{optimizers}. We set the \textit{dropout} to range between 0 and 0.95 and the \textit{momentum} between 0 and 0.9. Finally, we optimize the \textit{learning rate}, that is set to range between $10^{-4}$ and $10^{-1}$, on a logarithm scale, and the hyperparameter \textit{learn priors} that controls if the priors are learnable parameters. We fix the batch size to 200 and we adopted an early stopping criterion for determining the convergence of each model. Following~\cite{bianchiCTM}, we use the contextualized document representations derived from SentenceBERT~\cite{Reimers19sbert}. In particular, we use the pre-trained RoBERTa model fine-tuned on STS\footnote{\texttt{stsb-roberta-large}} for the English datasets and the multilingual Universal Sentence Encoder\footnote{\texttt{distiluse-base-multilingual-cased-v1}} for AFP.

For all the models, we set the remaining parameters to their default values. The models, metrics and the SOBO and MOBO pipelines are integrated into the open-source topic modeling framework OCTIS~\cite{Terragni21_octis,Terragni21_octis2}.\footnote{\url{https://github.com/Mind-lab/octis}}

\subsection{Hyperparameter Transfer Setting}
Regarding the hyperparameter transfer task, we consider the 5 best hyperparameter configurations for each metric, model, and dataset obtained during the multi-objective optimization experiments. We use the identified evaluations coming from a dataset to train a topic model on a different target dataset. As before, to obtain a robust result, we train the model with the same hyperparameter configuration 30 times and consider the median of the 30 evaluations. 

\section{Results}
\label{sec:results}
In the following, we discuss the results of the MOBO experiments and the hyperparameter transfer experiments. We report the best 5 hyperparameter configuration for each model and metric on each dataset in the Appendix.
\begin{figure*}[h]
\centering
         \includegraphics[width=\textwidth]{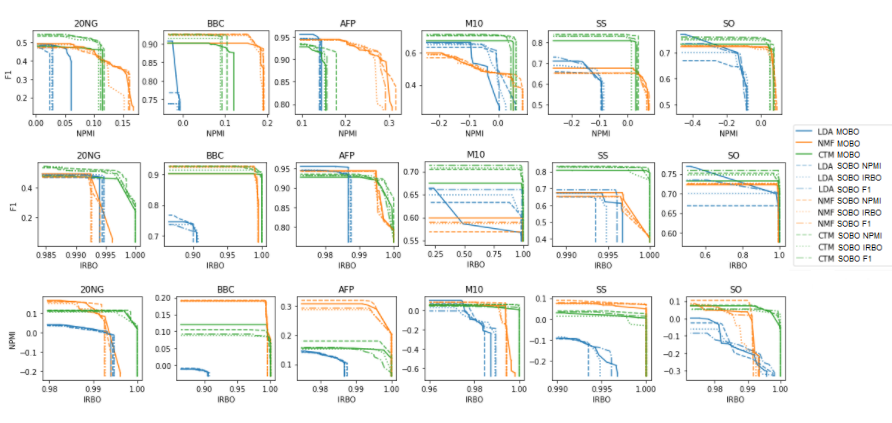}
         
     \caption{Pareto frontier for each pair of metrics on the considered datasets.}
     \label{fig:metricXmetric}
\end{figure*}

\subsection{Multi-objective hyperparameter optimization (RS \#1)}

\paragraph{No topic model wins them all.} Figure~\ref{fig:best_results} reports the best performance of the models for each metric and dataset obtained by the MOBO experiments. It is important to notice that the hyperparameter configuration that allows a topic model to obtain the best performance for a given metric may differ from the optimal hyperparameter configuration for another evaluation metric. 

Regarding the models' performance for the topic coherence (subfigure ~\ref{fig:best_results}.a), we can observe that NMF outperforms the other models in most cases. The stronger regularization in NMF generally leads to sparse topics and this likely leads to higher coherence scores~\cite{burkhardt2019decoupling}. 
Considering the predicting capabilities of the models (subfigure ~\ref{fig:best_results}.b), CTM usually outperforms LDA and NMF for short-text documents (M10, SS), while LDA gets the best results in long-text datasets (20NG, BBC, AFP). We note that CTM incorporates contextualized representations originated by a limited number of tokens, and not by the entire document. It follows that the representations of CTM may not produce accurate results for long-text documents. Finally, in the subfigure ~\ref{fig:best_results}.    c, we observe that CTM and NMF reach comparable topic diversity performance, often getting topics that are totally different from each other. 

On the same figure, we see the performance of the same models obtained by SOBO on each dataset. At first glance, we can see that for most combinations of dataset and metric, the MO models perform on par with their SO counterparts, suggesting that even under the multi-objective constraint, not much is sacrificed in performance on individual metrics (while improving the performance on the other metrics). 

As a concluding remark, except for the coherence, we showed that it is difficult to determine an always-winning topic model when we boost the performance of the models using multi-objective optimization. This finding is consistent with other investigations~\cite{Stevens12manymodels,Korencic18doccoherence}, despite that previous work did not optimize the models' hyperparameters. This result raises a question on the fairness of the past comparisons between topic models. This contributes to the growing amount of negative results when reviewing previously published work in light of new experiments~\cite{insights_negative_results}.

\paragraph{Conflicting objectives.} Although considering the best performance for each topic model can provide an indicator of its capabilities, it is essential not to focus on a single metric, but rather to jointly consider multiple objectives. Hereby, we show the trade-off between a pair of metrics by plotting the Pareto frontier of the considered metrics. 
Figure \ref{fig:metricXmetric} shows the frontier of each model for the pairs of metrics  (NPMI, IRBO), (F1, NPMI) and (F1, IRBO) respectively.  

We can observe that in most cases no model dominates the others, i.e. there is not any Pareto frontier that is better than the others for all the objectives. For example, if we consider the frontiers for NPMI and IRBO on 20NG, the frontier of the models CTM and NMF dominates LDA. However, CTM and NMF do not dominate each other. In other words, for the dataset 20NG, a user that aims at obtaining coherent and diverse topics has to compromise between the two objectives. 

In some cases, a topic model that outperforms the others for a given metric performs very poorly considering other metrics. For example, LDA on M10 obtains the best topic coherence but achieves a very low F1 (<0.2). Specifically, when considering F1 vs NPMI, we observe that to obtain high performance for a given metric we need to degrade the others, and vice versa. 

On the same figures, we can compare again this performance in between the single-objective and multi-objective. On most datasets, the MOBO models perform better than their SOBO counterparts. On the remaining datasets, we can see that the SOBO can perform better on singular metric it's trained on, thus pushing its Pareto frontier beyond that of its MOBO counterpart.

These results enforce the idea of not limiting the experimental campaign of topic models to a single-objective hyperparameter optimization approach. This approach has been suggested in~\cite{Terragni21_octis} for optimizing the hyperparameters using single-objective Bayesian Optimization, but it is also very common in other hyperparameter optimization settings (e.g. grid search). Such methodology may lead to poor or non-optimal results for the metrics that are not optimized. Yet, we should advocate for models that can guarantee the best trade-off among all the metrics of interest.

\begin{figure*}[h]
    \centering
    \includegraphics[width=1\textwidth]{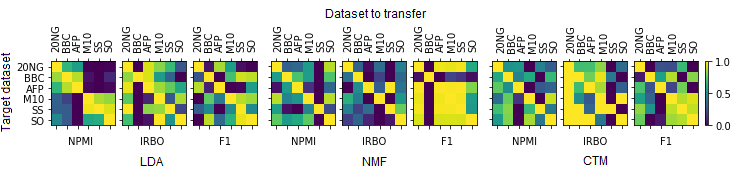}
    \caption{Heatmap matrix representing the performance when transferring the 5 best configurations from a dataset to a target dataset. The average of the runs is computed and each row is normalized between 0 and 1.
   }
    \label{fig:transfer_matrix}
\end{figure*}
\paragraph{The Cost of the Hyperparameter Optimization.}
Although optimizing the hyperparameters of a topic model guarantees a fair comparison with other models, this approach is computationally expensive, possibly making it difficult to replicate the results~\cite{bianchi2021gap}. In our work, we used BO because it is more efficient than other methods \cite{Snoek12practicalBO, Bergstra12}. Yet, the process requires a fair amount of iterations to guarantee the convergence to an optimal solution, especially when the hyperparameter space is large. Moreover, following \cite{Terragni21_octis}, we run the models with the same hyperparameter configuration for 30 times to guarantee robust results. It follows that replicating these results require time and computational resources. In the Appendix, we report an estimation of the cost the hyperparameter optimization.

In light of these observations, we argue that the knowledge that we have acquired for this extensive experimental campaign needs to be exploited and transferred. This will lead to results which are more accurate and obtained more efficiently. As previously mentioned, one direction is to use the best hyperparameter configurations on a dataset to initialize the hyperparameter optimization on the unseen target dataset~\cite{Feurer14metalearning}. In this paper, we do not discuss this direction in detail, but we will later show that hyperparameter transfer is effective in some cases and it is therefore a promising solution to minimize the computational cost to achieve optimal results on new datasets.

\subsection{Hyperparameter Transfer (RS \#2)}

\paragraph{Hyperparameter consistency across datasets.}

\begin{figure}[h]
     \centering
 \includegraphics[width=\columnwidth]{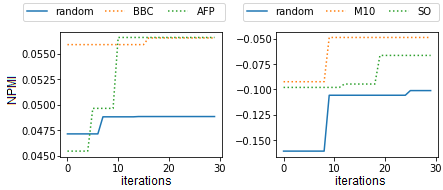}
         \caption{20NG dataset.}

     \label{fig:init_conf_ss}
     \caption{LDA performance in terms on NPMI with MOBO initialized with random configurations or with the configurations transferred from another dataset.}
\end{figure}

In the following, we report the results related to the hyperparameter transfer to an unseen dataset. This allows us to identify if the best hyperparameters are consistent across all the datasets. 
Figure \ref{fig:transfer_matrix} shows 
the results for LDA, CTM and NMF for each metric. Each matrix represents the performance of a model when the best 5 hyperparameter configurations are transferred from a dataset (columns) to the target dataset (rows). We compute the average of the 5 runs and we normalize each row. The diagonal of the matrix is then usually 1, since it represents the best configurations identified by the multi-objective optimization.
We report the disaggregated results in the Appendix. 

Let us consider the transfer on LDA (first three matrices on the left). 
Regarding NPMI, when we transfer the configurations from/to 20NG, AFP, and BBC, the topic model obtains results that are similar to the ones discovered by the previous MOBO experiments. Similar observations hold for the datasets M10, SS and SO. We can therefore deduce that the document length has an impact on the discovery of the best hyperparameter configuration for topic coherence. We observe a similar behavior for the IRBO performance, with the exception of the BBC dataset. Although the values of the topic diversity are very close to each other, the long-text datasets usually get similar performance when the hyperparameters are transferred from long-text datasets, and the same holds for short-text datasets. 

On the other hand, concerning the F1 performance, we observe a different trend: the configurations coming from BBC, M10, SS and SO seem to be transferable to that group of datasets, while a configuration coming from 20NG or AFP does not guarantee high performance on the previous datasets. This suggests that the document length is not the only feature that needs to be taken into consideration when we transfer the hyperparameters. 

Concerning the models NMF and CTM,  
we can observe different patterns for each metric. For example, if we consider the topic coherence in CTM, the configurations related to datasets 20NG, AFP, BBC and SS have close performances, but distant from M10 and SO. On the other hand, in NMF, the datasets 20NG, M10 and SO appear to be similar to each other, and distant from the others. This might be related to the fact that the topic models are regulated by different types of hyperparameters, which have a different impact on the models' objectives. 

In the considered experiments, we transfer the hyperparameters of a French dataset, i.e. AFP, to English datasets (and vice versa). The AFP's configurations transferred to another dataset can yield good results, 
potentially suggesting the existence of language pairs for which the parameter transfer can be near-optimal, given the similiraty of other dataset features. This result is extremely relevant because it would allow us to transfer the known best hyperparameter configurations in low-resource settings, when the best hyperparameter configuration for a dataset in a given language is not available or is expensive to compute. 

\paragraph{Random Initial Configurations vs Transferred Initial Configurations.} 
In this section, we will empirically show what is the effect of using a good set of initial configurations (obtained from the transfer of knowledge from previous experiments), compared to the random initialization. 

Figure  \ref{fig:init_conf_ss} shows
the NPMI performance of LDA for the first 30 MOBO iterations on 20NG (left) and SS (right), when the random initialization is performed (\textit{random}) or when the MOBO is initialized with the best configurations deriving from another dataset. We transfer the configurations from BBC and AFP for 20NG and the configurations from M10 and SO for SS, since these are the configurations that transferred better for 20NG and SS. Since the first 5 iterations are not ordered chronologically, we report the maximum of them. 

We can observe from both figures that, when using the transferred hyperparameters, the MOBO can achieve better results in just a few iterations, outperforming the 30 iterations of the MOBO initialized with random configurations. Therefore using the transferred configurations as initial ones can be helpful to obtain good results in less iterations. 





\section{Conclusion and Future Work}
\label{sec:conclusion}
In this paper, we investigated the role of a multi-objective optimization approach in topic models. We discovered that when we boost the models' performance at the best of their capabilities, it is not possible to identify an always-winning topic model for each considered objective, thus raising a question on the fairness of the past evaluations and comparisons between topic models. 
This result is further enforced when additional objectives are jointly considered. 
Thus, a user that aims to optimize multiple objectives at the same time has to compromise between the different objectives.  

We also showed that, in some cases, it is possible to effectively transfer the hyperparameters from a dataset to another. This result paves the way to exciting future research directions. In fact, the hyperparameter transfer allows researchers to avoid several and expensive iterations of hyperparameter optimization. In our experiments, we focused on the transfer of the best hyperparameters for a single metric. However, since we showed that some objectives are competing, a user may require to transfer a hyperparameter configuration that is a combination of more than one objective.

It is worth further investigating which dataset features contribute to a configuration transferable from a dataset to another. Our results suggest that the document length plays a role in the transfer, but other features such as the word and class distributions could be important too. We also showed that the dataset features are likely to be independent of the dataset language, leading to the use of hyperparameter transfer even to unseen datasets in different and low-resource languages.

\section*{Ethical Statement}
In this research work, we used datasets from the recent literature, and we do not use or infer any sensitive information. The risk of possible abuse of the models and proposed approach is low.

\bibliography{anthology,custom}
\bibliographystyle{acl_natbib}

\appendix

\section{Preprocessing details}
Datasets SS and SO are already preprocessed. The dataset source refers to \cite{phan08classification} for the dataset details; however the preprocessing pipeline is not available. According to \newcite{Terragni21_octis}, datasets 20NG, M10 and BBC have been preprocessed following these steps: tokenization, punctuation and number removal, lemmatization, stop-words removal (using the English stop-words list provided by MALLET), rare words removal and removal of short documents. In particular, they removed the words that have a word frequency less than 50\% for 20NG and BBC and less than 0.05\% for M10. And they removed the documents with less than 5 words for 20NG and BBC, and less than 3 words for M10. For APF, we followed the same preprocessing pipeline, except we removed the words that have document frequency lower than 1\% and higher than 70\% and we removed the documents with less than 5 words.

\section{The Cost of Hyperparameter Optimization}
In Table \ref{tab:costs} we report an average estimation of the time expressed in minutes for an iteration of the hyperparameter optimization for the considered models on the 20NG and M10. The overall running time of the optimization can vary depending on the number of documents, the dimensionality of the vocabulary, on the selected hyperparameters (e.g. the number of topics), and of course on the total iterations of the MOBO. 

\begin{table}[h]
\centering
\begin{tabular}{@{}lll@{}}
\toprule
 & \multicolumn{2}{l}{\textbf{Datasets}} \\ \toprule
 & {20NG} & {M10} \\\midrule
LDA & 37.51 & 16.45 \\
NMF & 42.16 & 21.66 \\
CTM & 65.98 & 39.24 \\ \bottomrule
\end{tabular}
\caption{Estimated minutes to complete one iteration of the MOBO for 20NG and M10 for each model.}
\label{tab:costs}
\end{table}
\section{Hyperparameter Transfer}
\begin{figure*}[]
    \centering
    \includegraphics[width=\textwidth]{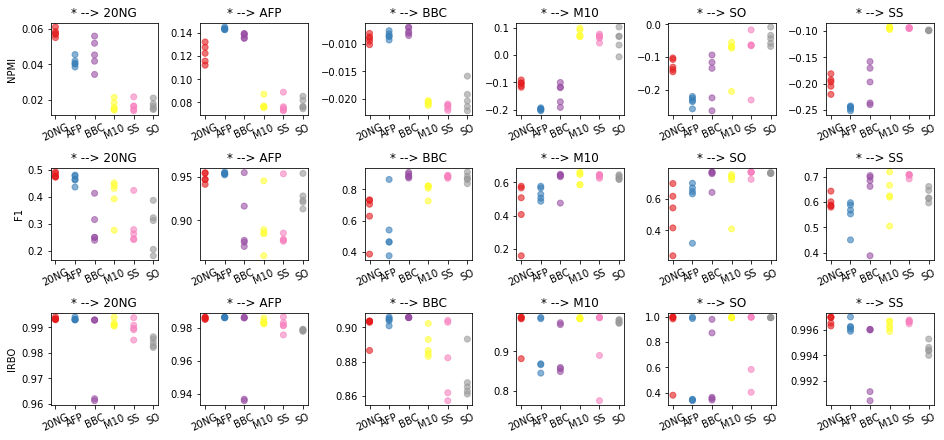}
    \caption{Training for LDA: * $\rightarrow d$ denotes that we transfer hyperparameters from the dataset * to train LDA on the dataset $d$. The x-axis reports the different datasets from which a configuration is transferred.}
    \label{fig:transfer_lda}
\end{figure*}

\begin{figure*}[]
    \centering
    \includegraphics[width=\textwidth]{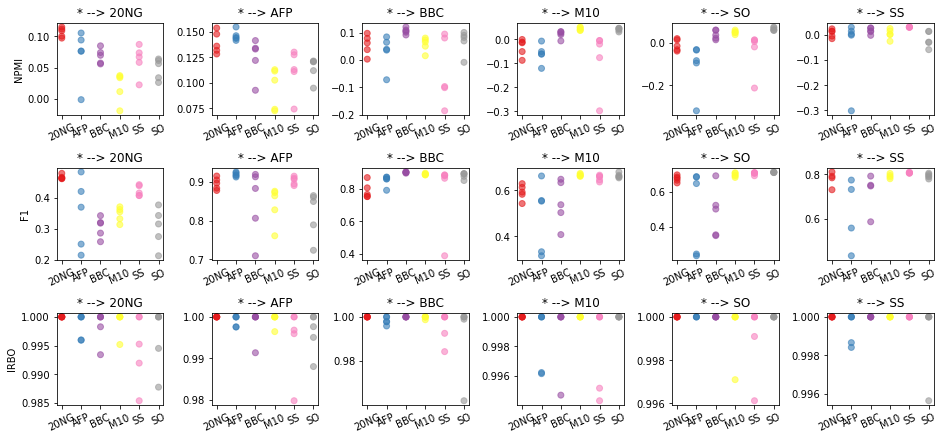}
    \caption{Training for CTM: * $\rightarrow d$ denotes that we transfer hyperparameters from the dataset * to train CTM on the dataset $d$. The x-axis reports the different datasets from which a configuration is transferred.}
    \label{fig:transfer_ctm}
\end{figure*}

\begin{figure*}[]
    \centering
    \includegraphics[width=\textwidth]{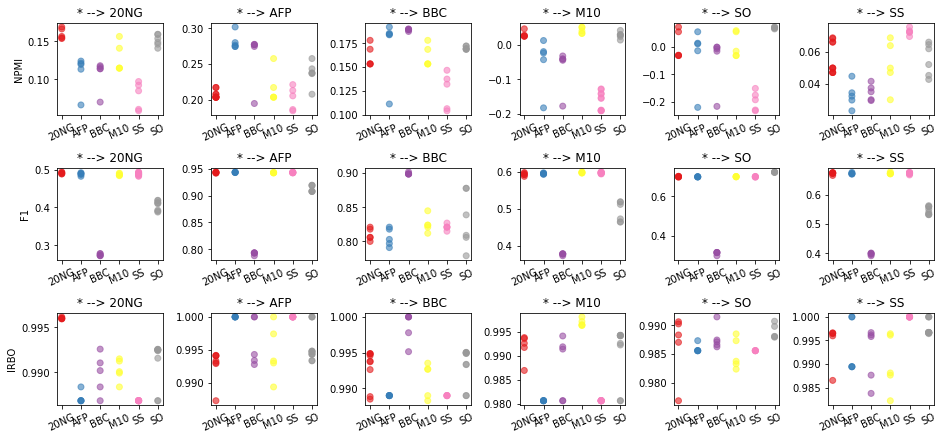}
    \caption{Training for NMF: * $\rightarrow d$ denotes that we transfer hyperparameters from the dataset * to train NMF on the dataset $d$. The x-axis reports the different datasets from which a configuration is transferred.}
    \label{fig:transfer_nmf}
\end{figure*}

Figures \ref{fig:transfer_lda}, \ref{fig:transfer_ctm} and \ref{fig:transfer_nmf} show reports the obtained value of the considered metric for the 5 best hyperparameter configurations that we transferred from a dataset (x-axis) to the target dataset ($\rightarrow$ dataset name). 


\section{Best hyperparameters configurations}
Tables \ref{tab:lda_npmi}, \ref{tab:lda_f1} and \ref{tab:lda_irbo} report the 5 best hyperparameter configurations for LDA for F1, NPMI, and IRBO respectively. Analogous details are provided in Tables \ref{tab:ctm_f1}, \ref{tab:ctm_npmi} and \ref{tab:ctm_irbo}, \ref{tab:nmf_f1}, \ref{tab:nmf_npmi} and \ref{tab:nmf_irbo} for CTM and NMF respectively.

\begin{table*}[h]
\centering
\resizebox{0.42\textwidth}{!}{%
\begin{tabular}{@{}crrrrrr@{}}
\toprule
Dataset & $\alpha$ prior & $\beta$ prior & \# Topics & IRBO & NPMI & F1 \\ \midrule
20NG & 0.0001 & 10.0000 & 20 & 0.95 & 0.061 & 0.397 \\
20NG & 0.0008 & 6.3062 & 26 & 0.953 & 0.058 & 0.424 \\
20NG & 0.0121 & 0.0001 & 21 & 0.95 & 0.057 & 0.408 \\
20NG & 0.0023 & 1.2349 & 27 & 0.961 & 0.057 & 0.428 \\
20NG & 0.0002 & 0.0002 & 19 & 0.946 & 0.055 & 0.403 \\\midrule
BBC & 0.1007 & 10.0000 & 32 & 0.827 & -0.007 & 0.73 \\
BBC & 0.0863 & 6.7863 & 48 & 0.85 & -0.007 & 0.742 \\
BBC & 0.1892 & 1.0099 & 66 & 0.861 & -0.008 & 0.746 \\
BBC & 0.0080 & 3.0815 & 38 & 0.836 & -0.008 & 0.751 \\
BBC & 0.0002 & 0.0001 & 46 & 0.849 & -0.009 & 0.627 \\\midrule
AFP & 0.0015 & 0.0001 & 46 & 0.976 & 0.145 & 0.942 \\
AFP & 0.0001 & 0.0001 & 49 & 0.976 & 0.144 & 0.94 \\
AFP & 0.0054 & 0.0008 & 44 & 0.976 & 0.143 & 0.941 \\
AFP & 0.0004 & 0.0010 & 50 & 0.977 & 0.143 & 0.942 \\
AFP & 0.1969 & 0.0002 & 46 & 0.977 & 0.143 & 0.944 \\\midrule
SS & 0.0001 & 4.8535 & 147 & 0.992 & -0.093 & 0.603 \\
SS & 0.0003 & 9.1500 & 149 & 0.991 & -0.094 & 0.538 \\
SS & 0.0119 & 10.0000 & 150 & 0.991 & -0.094 & 0.567 \\
SS & 0.0001 & 10.0000 & 150 & 0.991 & -0.095 & 0.518 \\
SS & 0.0001 & 10.0000 & 150 & 0.991 & -0.096 & 0.506 \\\midrule
M10 & 0.0030 & 10.0000 & 150 & 0.974 & 0.101 & 0.142 \\
M10 & 0.0013 & 10.0000 & 150 & 0.972 & 0.098 & 0.106 \\
M10 & 0.0024 & 10.0000 & 150 & 0.973 & 0.098 & 0.144 \\
M10 & 0.0011 & 10.0000 & 150 & 0.974 & 0.097 & 0.141 \\
M10 & 0.0014 & 10.0000 & 150 & 0.972 & 0.097 & 0.141 \\\midrule
SO & 0.0007 & 10.0000 & 133 & 0.97 & -0.008 & 0.093 \\
SO & 0.0019 & 10.0000 & 138 & 0.955 & -0.012 & 0.094 \\
SO & 0.0004 & 10.0000 & 138 & 0.955 & -0.024 & 0.093 \\
SO & 0.0004 & 10.0000 & 142 & 0.94 & -0.024 & 0.093 \\
SO & 0.0002 & 10.0000 & 142 & 0.941 & -0.025 & 0.093 \\ \bottomrule
\end{tabular}%
}
\caption{Best 5 hyperparameter configurations for LDA on each dataset for NPMI.}
\label{tab:lda_npmi}
\end{table*}

\begin{table*}[]
\centering
\resizebox{0.42\textwidth}{!}{%
\begin{tabular}{@{}crrrrrr@{}}
\toprule
Dataset & $\alpha$ prior & $\beta$ prior & \# Topics & IRBO & NPMI & F1 \\ \midrule
20NG & 0.0001 & 10.0000 & 20 & 0.95 & 0.061 & 0.397 \\
20NG & 0.0008 & 6.3062 & 26 & 0.953 & 0.058 & 0.424 \\
20NG & 0.0121 & 0.0001 & 21 & 0.95 & 0.057 & 0.408 \\
20NG & 0.0023 & 1.2349 & 27 & 0.961 & 0.057 & 0.428 \\
20NG & 0.0002 & 0.0002 & 19 & 0.946 & 0.055 & 0.403 \\\midrule
BBC & 0.1007 & 10.0000 & 32 & 0.827 & -0.007 & 0.73 \\
BBC & 0.0863 & 6.7863 & 48 & 0.85 & -0.007 & 0.742 \\
BBC & 0.1892 & 1.0099 & 66 & 0.861 & -0.008 & 0.746 \\
BBC & 0.0080 & 3.0815 & 38 & 0.836 & -0.008 & 0.751 \\
BBC & 0.0002 & 0.0001 & 46 & 0.849 & -0.009 & 0.627 \\\midrule
AFP & 0.0015 & 0.0001 & 46 & 0.976 & 0.145 & 0.942 \\
AFP & 0.0001 & 0.0001 & 49 & 0.976 & 0.144 & 0.94 \\
AFP & 0.0054 & 0.0008 & 44 & 0.976 & 0.143 & 0.941 \\
AFP & 0.0004 & 0.0010 & 50 & 0.977 & 0.143 & 0.942 \\
AFP & 0.1969 & 0.0002 & 46 & 0.977 & 0.143 & 0.944 \\\midrule
SS & 0.0001 & 4.8535 & 147 & 0.992 & -0.093 & 0.603 \\
SS & 0.0003 & 9.1500 & 149 & 0.991 & -0.094 & 0.538 \\
SS & 0.0119 & 10.0000 & 150 & 0.991 & -0.094 & 0.567 \\
SS & 0.0001 & 10.0000 & 150 & 0.991 & -0.095 & 0.518 \\
SS & 0.0001 & 10.0000 & 150 & 0.991 & -0.096 & 0.506 \\\midrule
M10 & 0.0030 & 10.0000 & 150 & 0.974 & 0.101 & 0.142 \\
M10 & 0.0013 & 10.0000 & 150 & 0.972 & 0.098 & 0.106 \\
M10 & 0.0024 & 10.0000 & 150 & 0.973 & 0.098 & 0.144 \\
M10 & 0.0011 & 10.0000 & 150 & 0.974 & 0.097 & 0.141 \\
M10 & 0.0014 & 10.0000 & 150 & 0.972 & 0.097 & 0.141 \\\midrule
SO & 0.0007 & 10.0000 & 133 & 0.97 & -0.008 & 0.093 \\
SO & 0.0019 & 10.0000 & 138 & 0.955 & -0.012 & 0.094 \\
SO & 0.0004 & 10.0000 & 138 & 0.955 & -0.024 & 0.093 \\
SO & 0.0004 & 10.0000 & 142 & 0.94 & -0.024 & 0.093 \\
SO & 0.0002 & 10.0000 & 142 & 0.941 & -0.025 & 0.093 \\ \bottomrule
\end{tabular}%
}
\caption{Best 5 hyperparameter configurations for LDA on each dataset for F1.}
\label{tab:lda_f1}
\end{table*}

\begin{table*}[]
\centering
\resizebox{0.42\textwidth}{!}{%
\begin{tabular}{@{}crrrrrr@{}}
\toprule
Dataset & $\alpha$ prior & $\beta$ prior & \# Topics & IRBO & NPMI & F1 \\ \midrule
20NG & 0.1483 & 0.0083 & 150 & 0.994 & -0.010 & 0.473 \\
20NG & 0.0446 & 0.0001 & 150 & 0.993 & -0.005 & 0.455 \\
20NG & 0.0947 & 0.0002 & 150 & 0.993 & 0.005 & 0.399 \\
20NG & 0.0048 & 0.2006 & 150 & 0.993 & -0.003 & 0.434 \\
20NG & 0.1961 & 0.0001 & 111 & 0.993 & 0.010 & 0.365 \\\midrule
BBC & 0.0001 & 0.0001 & 150 & 0.906 & -0.021 & 0.357 \\
BBC & 0.0447 & 10.0000 & 150 & 0.906 & -0.020 & 0.691 \\
BBC & 0.0001 & 0.0001 & 150 & 0.906 & -0.021 & 0.348 \\
BBC & 0.0004 & 10.0000 & 150 & 0.906 & -0.020 & 0.697 \\
BBC & 0.0003 & 0.0071 & 150 & 0.906 & -0.020 & 0.325 \\\midrule
AFP & 0.1668 & 0.0002 & 150 & 0.987 & 0.103 & 0.953 \\
AFP & 0.0452 & 0.0001 & 148 & 0.987 & 0.107 & 0.948 \\
AFP & 0.0017 & 0.0001 & 149 & 0.986 & 0.107 & 0.943 \\
AFP & 0.0006 & 0.0015 & 150 & 0.986 & 0.107 & 0.943 \\
AFP & 0.0001 & 0.0001 & 150 & 0.986 & 0.107 & 0.943 \\\midrule
SS & 0.0990 & 0.4641 & 98 & 0.997 & -0.256 & 0.557 \\
SS & 0.1242 & 0.0014 & 150 & 0.997 & -0.273 & 0.577 \\
SS & 0.2860 & 0.0085 & 69 & 0.997 & -0.265 & 0.579 \\
SS & 0.3039 & 0.9328 & 83 & 0.997 & -0.220 & 0.611 \\
SS & 0.0052 & 0.3995 & 149 & 0.996 & -0.236 & 0.553 \\\midrule
M10 & 0.1483 & 0.0001 & 150 & 0.987 & -0.236 & 0.291 \\
M10 & 0.1496 & 0.0187 & 103 & 0.985 & -0.252 & 0.556 \\
M10 & 0.0524 & 0.2615 & 102 & 0.985 & -0.239 & 0.429 \\
M10 & 0.0130 & 0.0011 & 105 & 0.984 & -0.209 & 0.500 \\
M10 & 0.0249 & 0.0001 & 117 & 0.984 & -0.207 & 0.510 \\\midrule
SO & 0.1657 & 0.0001 & 64 & 0.996 & -0.302 & 0.585 \\
SO & 0.0863 & 0.5342 & 81 & 0.995 & -0.297 & 0.605 \\
SO & 0.0808 & 0.0037 & 92 & 0.993 & -0.304 & 0.607 \\
SO & 0.0135 & 0.4842 & 68 & 0.993 & -0.264 & 0.503 \\
SO & 0.0223 & 0.5298 & 62 & 0.993 & -0.262 & 0.504 \\ \hline
\end{tabular}%
}
\caption{Best 5 hyperparameter configurations for LDA on each dataset for IRBO.}
\label{tab:lda_irbo}
\end{table*}

\begin{table*}[]
\resizebox{\textwidth}{!}{%
\begin{tabular}{clrcrrrrrlrrr}
\toprule
Dataset & Activation & Dropout & Learn priors & Learning rate & Momentum & \# Layers & \# Topics & \# Neurons & Optimizer & IRBO & NPMI & F1 \\\midrule
20NG & rrelu & 0.000 & 0 & 0.0001 & 0.114 & 1 & 143 & 1000 & rmsprop & 0.989 & 0.024 & 0.480 \\
20NG & elu & 0.113 & 1 & 0.0020 & 0.489 & 2 & 53 & 600 & adam & 0.994 & 0.094 & 0.467 \\
20NG & elu & 0.061 & 1 & 0.0001 & 0.387 & 2 & 118 & 100 & rmsprop & 0.985 & 0.044 & 0.464 \\
20NG & leakyrelu & 0.326 & 1 & 0.1000 & 0.808 & 1 & 63 & 400 & rmsprop & 0.995 & 0.082 & 0.464 \\
20NG & rrelu & 0.022 & 0 & 0.1000 & 0.200 & 1 & 64 & 200 & adam & 0.996 & 0.097 & 0.462 \\\midrule
AFP & leakyrelu & 0.274 & 1 & 0.0626 & 0.116 & 4 & 148 & 1000 & rmsprop & 0.989 & 0.104 & 0.926 \\
AFP & rrelu & 0.322 & 1 & 0.0015 & 0.900 & 1 & 150 & 1000 & rmsprop & 0.993 & 0.101 & 0.923 \\
AFP & rrelu & 0.354 & 0 & 0.0440 & 0.900 & 3 & 126 & 600 & rmsprop & 0.988 & 0.113 & 0.919 \\
AFP & rrelu & 0.115 & 1 & 0.0207 & 0.018 & 4 & 138 & 300 & adam & 0.991 & 0.112 & 0.917 \\
AFP & softplus & 0.000 & 0 & 0.1000 & 0.755 & 2 & 61 & 900 & adam & 0.995 & 0.145 & 0.913 \\\midrule
BBC & rrelu & 0.312 & 0 & 0.0149 & 0.420 & 3 & 139 & 900 & adam & 0.988 & 0.000 & 0.901 \\
BBC & selu & 0.327 & 0 & 0.0025 & 0.720 & 3 & 150 & 600 & adam & 0.978 & 0.062 & 0.901 \\
BBC & selu & 0.000 & 1 & 0.0055 & 0.061 & 1 & 9 & 800 & rmsprop & 1.000 & 0.066 & 0.899 \\
BBC & elu & 0.606 & 0 & 0.0001 & 0.482 & 2 & 78 & 300 & sgd & 0.948 & 0.075 & 0.899 \\
BBC & rrelu & 0.220 & 1 & 0.0013 & 0.900 & 1 & 6 & 900 & rmsprop & 1.000 & 0.004 & 0.894 \\\midrule
M10 & elu & 0.438 & 0 & 0.0012 & 0.012 & 1 & 16 & 1000 & sgd & 0.985 & 0.042 & 0.674 \\
M10 & selu & 0.645 & 0 & 0.0006 & 0.748 & 1 & 15 & 800 & sgd & 0.973 & 0.048 & 0.670 \\
M10 & elu & 0.549 & 0 & 0.0025 & 0.012 & 1 & 24 & 300 & rmsprop & 0.980 & 0.026 & 0.665 \\
M10 & elu & 0.708 & 1 & 0.0066 & 0.363 & 1 & 37 & 300 & adam & 0.971 & 0.023 & 0.664 \\
M10 & elu & 0.640 & 1 & 0.0003 & 0.304 & 1 & 37 & 100 & sgd & 0.964 & 0.051 & 0.661 \\\midrule
SO & elu & 0.367 & 1 & 0.0005 & 0.558 & 3 & 22 & 600 & adam & 0.990 & 0.042 & 0.732 \\
SO & selu & 0.126 & 1 & 0.0004 & 0.377 & 1 & 44 & 700 & adam & 0.987 & -0.019 & 0.721 \\
SO & elu & 0.577 & 1 & 0.1000 & 0.134 & 3 & 26 & 400 & adadelta & 0.972 & 0.034 & 0.717 \\
SO & sigmoid & 0.727 & 1 & 0.0009 & 0.716 & 1 & 23 & 600 & adam & 0.974 & 0.050 & 0.715 \\
SO & elu & 0.000 & 0 & 0.0001 & 0.840 & 2 & 19 & 600 & rmsprop & 0.996 & 0.043 & 0.714 \\\midrule
SS & elu & 0.452 & 1 & 0.0027 & 0.285 & 2 & 44 & 300 & adam & 0.991 & 0.011 & 0.809 \\
SS & selu & 0.777 & 1 & 0.0005 & 0.590 & 1 & 127 & 400 & adam & 0.977 & 0.023 & 0.807 \\
SS & leakyrelu & 0.034 & 1 & 0.0721 & 0.054 & 2 & 26 & 800 & rmsprop & 1.000 & -0.009 & 0.806 \\
SS & selu & 0.489 & 1 & 0.0057 & 0.084 & 2 & 59 & 800 & rmsprop & 0.992 & -0.024 & 0.805 \\
SS & rrelu & 0.562 & 1 & 0.0014 & 0.794 & 1 & 71 & 300 & rmsprop & 0.990 & -0.012 & 0.804\\\bottomrule
\end{tabular}%
}
\caption{Best 5 hyperparameter configurations for CTM on each dataset for F1.}
\label{tab:ctm_f1}
\end{table*}

\begin{table*}[]
\resizebox{\textwidth}{!}{%
\begin{tabular}{clrcrrrrrlrrr}
\toprule
Dataset & Activation & Dropout & Learn priors & Learning rate & Momentum & \# Layers & \# Topics & \# Neurons & Optimizer & IRBO & NPMI & F1 \\\midrule
20NG & elu & 0.000 & 0 & 0.1000 & 0.859 & 1 & 27 & 1000 & rmsprop & 0.997 & 0.115 & 0.461 \\
20NG & elu & 0.127 & 0 & 0.0016 & 0.009 & 1 & 30 & 900 & rmsprop & 0.996 & 0.112 & 0.458 \\
20NG & leakyrelu & 0.000 & 1 & 0.0052 & 0.650 & 2 & 44 & 600 & rmsprop & 0.995 & 0.109 & 0.456 \\
20NG & elu & 0.000 & 1 & 0.0712 & 0.097 & 3 & 49 & 1000 & adam & 0.995 & 0.100 & 0.424 \\
20NG & leakyrelu & 0.356 & 0 & 0.0028 & 0.744 & 1 & 22 & 1000 & rmsprop & 0.995 & 0.097 & 0.409 \\\midrule
BBC & softplus & 0.000 & 1 & 0.0018 & 0.306 & 1 & 18 & 400 & adam & 0.994 & 0.120 & 0.876 \\
BBC & softplus & 0.272 & 0 & 0.0002 & 0.066 & 1 & 16 & 200 & sgd & 0.977 & 0.109 & 0.878 \\
BBC & relu & 0.003 & 1 & 0.0002 & 0.067 & 1 & 37 & 700 & sgd & 0.977 & 0.106 & 0.797 \\
BBC & elu & 0.000 & 0 & 0.0001 & 0.500 & 3 & 10 & 300 & sgd & 0.984 & 0.101 & 0.890 \\
BBC & elu & 0.130 & 0 & 0.0001 & 0.453 & 4 & 26 & 200 & sgd & 0.972 & 0.091 & 0.890 \\\midrule
AFP & rrelu & 0.000 & 0 & 0.1000 & 0.105 & 4 & 129 & 900 & adagrad & 0.989 & 0.155 & 0.887 \\
AFP & rrelu & 0.000 & 0 & 0.0495 & 0.089 & 3 & 41 & 100 & rmsprop & 0.995 & 0.146 & 0.898 \\
AFP & softplus & 0.000 & 0 & 0.1000 & 0.755 & 2 & 61 & 900 & adam & 0.995 & 0.145 & 0.913 \\
AFP & relu & 0.000 & 0 & 0.0075 & 0.701 & 1 & 97 & 1000 & rmsprop & 0.994 & 0.144 & 0.896 \\
AFP & relu & 0.000 & 1 & 0.0875 & 0.856 & 4 & 58 & 600 & adam & 0.995 & 0.142 & 0.911 \\\midrule
SS & selu & 0.802 & 1 & 0.0001 & 0.781 & 3 & 38 & 100 & rmsprop & 0.952 & 0.031 & 0.757 \\
SS & relu & 0.296 & 0 & 0.0001 & 0.640 & 1 & 42 & 100 & rmsprop & 0.981 & 0.031 & 0.773 \\
SS & selu & 0.360 & 0 & 0.0002 & 0.688 & 1 & 42 & 700 & rmsprop & 0.990 & 0.031 & 0.801 \\
SS & selu & 0.359 & 0 & 0.0001 & 0.643 & 1 & 34 & 800 & rmsprop & 0.991 & 0.030 & 0.802 \\
SS & selu & 0.824 & 1 & 0.0002 & 0.469 & 1 & 105 & 400 & rmsprop & 0.969 & 0.029 & 0.802 \\\midrule
M10 & elu & 0.640 & 1 & 0.0003 & 0.304 & 1 & 37 & 100 & sgd & 0.964 & 0.051 & 0.661 \\
M10 & selu & 0.645 & 0 & 0.0006 & 0.748 & 1 & 15 & 800 & sgd & 0.973 & 0.048 & 0.670 \\
M10 & elu & 0.438 & 0 & 0.0012 & 0.012 & 1 & 16 & 1000 & sgd & 0.985 & 0.042 & 0.674 \\
M10 & leakyrelu & 0.393 & 0 & 0.0005 & 0.111 & 2 & 22 & 700 & sgd & 0.972 & 0.038 & 0.648 \\
M10 & softplus & 0.300 & 1 & 0.0006 & 0.694 & 2 & 30 & 800 & sgd & 0.974 & 0.036 & 0.657 \\\midrule
SO & sigmoid & 0.013 & 1 & 0.0016 & 0.442 & 1 & 18 & 100 & sgd & 0.991 & 0.073 & 0.701 \\
SO & selu & 0.000 & 1 & 0.0023 & 0.130 & 2 & 18 & 300 & sgd & 0.992 & 0.070 & 0.712 \\
SO & relu & 0.000 & 1 & 0.0005 & 0.870 & 2 & 16 & 1000 & sgd & 0.993 & 0.062 & 0.679 \\
SO & rrelu & 0.041 & 0 & 0.0050 & 0.305 & 2 & 19 & 200 & sgd & 0.990 & 0.060 & 0.690 \\
SO & leakyrelu & 0.482 & 1 & 0.0004 & 0.154 & 1 & 17 & 800 & sgd & 0.987 & 0.058 & 0.698 \\ \bottomrule
\end{tabular}%
}
\caption{Best 5 hyperparameter configurations for CTM on each dataset for NPMI.}
\label{tab:ctm_npmi}
\end{table*}

\begin{table*}[]
\resizebox{\textwidth}{!}{%
\begin{tabular}{@{}clrcrrrrrlrrr@{}}
\toprule
Dataset & Activation & Dropout & Learn priors & Learning rate & Momentum & \# Layers & \# Topics & \# Neurons & Optimizer & IRBO & NPMI & F1 \\\midrule
20NG & rrelu & 0.0238 & 0 & 0.0078 & 0.741 & 3 & 5 & 400 & rmsprop & 1.000 & 0.021 & 0.215 \\
20NG & softplus & 0.4102 & 1 & 0.0580 & 0.270 & 2 & 5 & 600 & rmsprop & 1.000 & 0.013 & 0.203 \\
20NG & rrelu & 0.0000 & 1 & 0.0874 & 0.389 & 1 & 5 & 400 & adagrad & 1.000 & 0.004 & 0.235 \\
20NG & rrelu & 0.0402 & 0 & 0.1000 & 0.552 & 4 & 5 & 100 & adagrad & 1.000 & -0.006 & 0.197 \\
20NG & leakyrelu & 0.0000 & 1 & 0.0633 & 0.385 & 1 & 5 & 500 & adam & 1.000 & 0.016 & 0.242 \\\midrule
BBC & rrelu & 0.0225 & 1 & 0.0676 & 0.864 & 4 & 5 & 900 & adam & 1.000 & -0.039 & 0.843 \\
BBC & selu & 0.0000 & 1 & 0.0055 & 0.061 & 1 & 9 & 800 & rmsprop & 1.000 & 0.066 & 0.899 \\
BBC & rrelu & 0.2199 & 1 & 0.0013 & 0.900 & 1 & 6 & 900 & rmsprop & 1.000 & 0.004 & 0.894 \\
BBC & relu & 0.0000 & 0 & 0.0005 & 0.038 & 1 & 5 & 700 & adadelta & 1.000 & -0.433 & 0.342 \\
BBC & rrelu & 0.0177 & 1 & 0.0792 & 0.819 & 1 & 5 & 900 & adagrad & 1.000 & -0.032 & 0.894 \\\midrule
AFP & selu & 0.6558 & 1 & 0.0060 & 0.540 & 1 & 5 & 600 & rmsprop & 1.000 & 0.051 & 0.657 \\
AFP & leakyrelu & 0.0000 & 1 & 0.0712 & 0.439 & 3 & 5 & 800 & adagrad & 1.000 & 0.033 & 0.667 \\
AFP & rrelu & 0.0000 & 1 & 0.1000 & 0.763 & 4 & 5 & 200 & adagrad & 1.000 & 0.025 & 0.665 \\
AFP & leakyrelu & 0.0000 & 0 & 0.0009 & 0.532 & 1 & 5 & 400 & rmsprop & 1.000 & 0.064 & 0.666 \\
AFP & rrelu & 0.0000 & 1 & 0.0712 & 0.807 & 2 & 8 & 500 & adagrad & 1.000 & 0.083 & 0.781 \\\midrule
SS & rrelu & 0.0000 & 1 & 0.0578 & 0.455 & 4 & 17 & 700 & rmsprop & 1.000 & -0.030 & 0.794 \\
SS & selu & 0.0000 & 1 & 0.0013 & 0.900 & 5 & 5 & 800 & adam & 1.000 & -0.105 & 0.549 \\
SS & selu & 0.0000 & 1 & 0.0006 & 0.729 & 1 & 7 & 200 & adam & 1.000 & 0.005 & 0.694 \\
SS & relu & 0.1055 & 1 & 0.0088 & 0.836 & 4 & 5 & 300 & adam & 1.000 & -0.155 & 0.530 \\
SS & relu & 0.8341 & 0 & 0.0010 & 0.900 & 2 & 5 & 1000 & adam & 1.000 & -0.153 & 0.523 \\\midrule
M10 & rrelu & 0.0000 & 0 & 0.0086 & 0.791 & 1 & 5 & 700 & adam & 1.000 & -0.083 & 0.496 \\
M10 & leakyrelu & 0.0329 & 0 & 0.0059 & 0.214 & 2 & 5 & 700 & rmsprop & 1.000 & -0.101 & 0.478 \\
M10 & selu & 0.0000 & 1 & 0.0196 & 0.088 & 1 & 5 & 300 & adam & 1.000 & -0.116 & 0.490 \\
M10 & relu & 0.0000 & 1 & 0.0489 & 0.157 & 1 & 5 & 400 & adam & 1.000 & -0.134 & 0.492 \\
M10 & elu & 0.6812 & 1 & 0.1000 & 0.013 & 1 & 6 & 300 & adam & 1.000 & -0.073 & 0.520 \\\midrule
SO & softplus & 0.3098 & 1 & 0.0001 & 0.527 & 3 & 5 & 600 & rmsprop & 1.000 & -0.135 & 0.304 \\\
SO & rrelu & 0.4490 & 1 & 0.0167 & 0.602 & 1 & 5 & 300 & rmsprop & 1.000 & -0.162 & 0.305 \\
SO & selu & 0.5117 & 0 & 0.0233 & 0.752 & 1 & 5 & 400 & rmsprop & 1.000 & -0.147 & 0.310 \\
SO & elu & 0.1316 & 1 & 0.0006 & 0.405 & 2 & 5 & 900 & sgd & 1.000 & -0.052 & 0.290 \\
SO & softplus & 0.2041 & 1 & 0.0002 & 0.312 & 3 & 5 & 1000 & rmsprop & 1.000 & -0.116 & 0.308 \\ \bottomrule
\end{tabular}%
}
\caption{Best 5 hyperparameter configurations for CTM on each dataset for IRBO.}
\label{tab:ctm_irbo}
\end{table*}

\begin{table*}[]
\centering
\resizebox{0.68\textwidth}{!}{%
\begin{tabular}{@{}crrllrrrr@{}}
\toprule
Dataset & Reg. factor & L1/L2 & Initialization & Regularization & \# Topics & IRBO & NPMI & F1 \\ \midrule
20NG & 0.000 & 1.000 & random & H matrix & 150 & 0.993 & 0.060 & 0.494 \\
20NG & 0.109 & 0.110 & random & H matrix & 150 & 0.992 & 0.059 & 0.492 \\
20NG & 0.000 & 1.000 & nndsvda & H matrix & 150 & 0.992 & 0.061 & 0.491 \\
20NG & 0.500 & 0.000 & nndsvda & V matrix & 150 & 0.992 & 0.061 & 0.489 \\
20NG & 0.001 & 0.537 & random & H matrix & 140 & 0.992 & 0.064 & 0.489 \\\midrule
BBC & 0.336 & 0.000 & nndsvdar & both & 5 & 0.989 & 0.153 & 0.901 \\
BBC & 0.068 & 0.156 & nndsvd & V matrix & 5 & 0.989 & 0.153 & 0.899 \\
BBC & 0.000 & 0.738 & nndsvda & both & 5 & 0.989 & 0.153 & 0.899 \\
BBC & 0.000 & 0.000 & nndsvda & both & 5 & 0.989 & 0.153 & 0.899 \\
BBC & 0.487 & 0.000 & random & V matrix & 5 & 0.989 & 0.153 & 0.899 \\\midrule
AFP & 0.131 & 0.000 & random & both & 150 & 0.994 & 0.188 & 0.944 \\
AFP & 0.139 & 0.280 & random & H matrix & 147 & 0.994 & 0.186 & 0.944 \\
AFP & 0.050 & 0.991 & random & H matrix & 150 & 0.994 & 0.184 & 0.943 \\
AFP & 0.314 & 0.000 & nndsvdar & H matrix & 150 & 0.995 & 0.186 & 0.943 \\
AFP & 0.445 & 0.066 & nndsvdar & H matrix & 148 & 0.995 & 0.185 & 0.943 \\\midrule
SS & 0.500 & 0.507 & nndsvda & V matrix & 150 & 0.997 & 0.022 & 0.676 \\
SS & 0.000 & 0.174 & nndsvd & both & 146 & 0.997 & 0.019 & 0.674 \\
SS & 0.000 & 0.021 & nndsvdar & H matrix & 150 & 0.997 & 0.017 & 0.674 \\
SS & 0.015 & 0.000 & nndsvdar & both & 150 & 0.997 & 0.018 & 0.674 \\
SS & 0.219 & 0.000 & nndsvda & V matrix & 150 & 0.997 & 0.018 & 0.673 \\\midrule
M10 & 0.275 & 0.000 & nndsvda & H matrix & 150 & 0.994 & -0.191 & 0.599 \\
M10 & 0.000 & 0.408 & nndsvdar & both & 150 & 0.994 & -0.192 & 0.598 \\
M10 & 0.477 & 0.000 & nndsvdar & V matrix & 150 & 0.994 & -0.191 & 0.596 \\
M10 & 0.025 & 0.534 & nndsvd & V matrix & 150 & 0.994 & -0.192 & 0.596 \\
M10 & 0.000 & 0.696 & nndsvdar & V matrix & 146 & 0.994 & -0.189 & 0.596 \\\midrule
SO & 0.390 & 0.570 & nndsvda & V matrix & 21 & 0.977 & 0.034 & 0.722 \\
SO & 0.428 & 0.563 & nndsvdar & H matrix & 46 & 0.986 & -0.075 & 0.721 \\
SO & 0.130 & 0.739 & nndsvd & H matrix & 43 & 0.986 & -0.067 & 0.721 \\
SO & 0.140 & 0.087 & random & V matrix & 44 & 0.986 & -0.070 & 0.721 \\
SO & 0.395 & 0.730 & nndsvdar & V matrix & 25 & 0.981 & -0.004 & 0.720 \\ \bottomrule
\end{tabular}%
}
\caption{Best 5 hyperparameter configurations for NMF on each dataset for F1.}
\label{tab:nmf_f1}
\end{table*}

\begin{table*}[]
\centering
\resizebox{0.68\textwidth}{!}{%
\begin{tabular}{@{}crrllrrrr@{}}
\toprule
Dataset & Reg. factor & L1/L2 & Initialization & Regularization & \# Topics & IRBO & NPMI & F1 \\ \midrule
20NG & 0.500 & 0.510 & nndsvdar & both & 5 & 0.983 & 0.169 & 0.150 \\
20NG & 0.423 & 0.555 & nndsvdar & both & 5 & 0.984 & 0.166 & 0.156 \\
20NG & 0.409 & 0.303 & nndsvda & V matrix & 9 & 0.988 & 0.156 & 0.325 \\
20NG & 0.312 & 0.350 & nndsvda & both & 5 & 0.980 & 0.154 & 0.208 \\
20NG & 0.026 & 0.395 & nndsvda & H matrix & 10 & 0.987 & 0.154 & 0.339 \\\midrule
BBC & 0.500 & 0.788 & nndsvda & H matrix & 28 & 0.992 & 0.190 & 0.833 \\
BBC & 0.429 & 0.344 & nndsvd & both & 125 & 0.959 & 0.190 & 0.487 \\
BBC & 0.234 & 0.264 & nndsvd & V matrix & 26 & 0.992 & 0.189 & 0.818 \\
BBC & 0.486 & 0.000 & nndsvd & H matrix & 26 & 0.992 & 0.189 & 0.830 \\
BBC & 0.465 & 0.433 & nndsvda & H matrix & 27 & 0.992 & 0.188 & 0.821 \\\midrule
AFP & 0.415 & 0.788 & nndsvdar & both & 133 & 0.993 & 0.302 & 0.806 \\
AFP & 0.426 & 0.804 & nndsvdar & H matrix & 20 & 0.994 & 0.280 & 0.900 \\
AFP & 0.114 & 0.078 & random & V matrix & 24 & 0.994 & 0.276 & 0.906 \\
AFP & 0.000 & 0.705 & nndsvdar & both & 28 & 0.994 & 0.275 & 0.904 \\
AFP & 0.000 & 0.000 & random & V matrix & 24 & 0.994 & 0.274 & 0.907 \\\midrule
SS & 0.379 & 0.283 & nndsvd & both & 146 & 0.994 & 0.076 & 0.512 \\
SS & 0.179 & 0.708 & nndsvdar & both & 99 & 0.993 & 0.073 & 0.543 \\
SS & 0.293 & 0.255 & nndsvdar & both & 87 & 0.995 & 0.073 & 0.570 \\
SS & 0.205 & 0.477 & nndsvd & both & 150 & 0.994 & 0.072 & 0.548 \\
SS & 0.248 & 0.319 & nndsvdar & both & 75 & 0.995 & 0.070 & 0.563 \\\midrule
M10 & 0.000 & 0.025 & random & V matrix & 5 & 0.962 & 0.051 & 0.362 \\
M10 & 0.037 & 1.000 & nndsvdar & V matrix & 9 & 0.994 & 0.049 & 0.430 \\
M10 & 0.120 & 0.854 & nndsvd & H matrix & 15 & 0.989 & 0.039 & 0.468 \\
M10 & 0.031 & 0.605 & nndsvd & V matrix & 5 & 0.981 & 0.032 & 0.368 \\
M10 & 0.080 & 1.000 & nndsvdar & H matrix & 5 & 0.981 & 0.032 & 0.369 \\\midrule
SO & 0.443 & 0.741 & nndsvdar & V matrix & 12 & 0.974 & 0.071 & 0.578 \\
SO & 0.090 & 0.673 & nndsvdar & both & 10 & 0.975 & 0.070 & 0.484 \\
SO & 0.182 & 0.720 & nndsvd & V matrix & 14 & 0.975 & 0.068 & 0.604 \\
SO & 0.440 & 0.612 & nndsvdar & H matrix & 13 & 0.974 & 0.066 & 0.587 \\
SO & 0.018 & 0.378 & nndsvdar & V matrix & 15 & 0.973 & 0.064 & 0.628 \\ \bottomrule
\end{tabular}%
}
\caption{Best 5 hyperparameter configurations for NMF on each dataset for NPMI.}
\label{tab:nmf_npmi}
\end{table*}

\begin{table*}[]
\centering
\resizebox{0.68\textwidth}{!}{%
\begin{tabular}{@{}crrllrrrr@{}}
\toprule
Dataset & Reg. factor & L1/L2 & Initialization & Regularization & \# Topics & IRBO & NPMI & F1 \\ \midrule
20NG & 0.250 & 1.000 & random & V matrix & 11 & 0.996 & -0.207 & 0.056 \\
20NG & 0.443 & 0.980 & random & V matrix & 87 & 0.996 & -0.195 & 0.056 \\
20NG & 0.479 & 0.607 & random & V matrix & 124 & 0.996 & -0.193 & 0.056 \\
20NG & 0.204 & 0.795 & random & V matrix & 131 & 0.996 & -0.195 & 0.056 \\
20NG & 0.500 & 0.919 & random & V matrix & 67 & 0.996 & -0.196 & 0.056 \\\midrule
BBC & 0.390 & 0.786 & nndsvd & both & 68 & 1.000 & 0.077 & 0.301 \\
BBC & 0.281 & 0.998 & random & V matrix & 5 & 1.000 & -0.431 & 0.227 \\
BBC & 0.500 & 1.000 & random & both & 5 & 1.000 & -0.431 & 0.227 \\
BBC & 0.500 & 0.478 & random & V matrix & 57 & 0.998 & -0.416 & 0.227 \\
BBC & 0.142 & 0.447 & random & H matrix & 150 & 0.995 & 0.101 & 0.803 \\\midrule
AFP & 0.335 & 0.000 & random & H matrix & 5 & 1.000 & 0.203 & 0.793 \\
AFP & 0.372 & 1.000 & nndsvd & V matrix & 5 & 1.000 & 0.203 & 0.788 \\
AFP & 0.366 & 0.883 & nndsvda & V matrix & 5 & 1.000 & 0.203 & 0.788 \\
AFP & 0.282 & 1.000 & random & H matrix & 5 & 1.000 & 0.197 & 0.791 \\
AFP & 0.011 & 0.000 & random & H matrix & 5 & 1.000 & 0.203 & 0.793 \\\midrule
SS & 0.072 & 0.018 & nndsvda & V matrix & 5 & 1.000 & 0.050 & 0.400 \\
SS & 0.000 & 0.742 & nndsvdar & V matrix & 5 & 1.000 & 0.050 & 0.401 \\
SS & 0.000 & 0.849 & nndsvda & H matrix & 5 & 1.000 & 0.050 & 0.395 \\
SS & 0.199 & 1.000 & nndsvdar & V matrix & 5 & 1.000 & 0.047 & 0.396 \\
SS & 0.323 & 1.000 & nndsvd & V matrix & 5 & 1.000 & 0.047 & 0.392 \\\midrule
M10 & 0.500 & 0.182 & random & V matrix & 5 & 0.998 & -0.643 & 0.135 \\
M10 & 0.215 & 0.804 & random & V matrix & 6 & 0.997 & -0.633 & 0.135 \\
M10 & 0.430 & 0.736 & random & V matrix & 7 & 0.996 & -0.624 & 0.135 \\
M10 & 0.092 & 0.264 & random & V matrix & 86 & 0.996 & -0.514 & 0.135 \\
M10 & 0.140 & 0.789 & random & V matrix & 70 & 0.996 & -0.516 & 0.135 \\\midrule
SO & 0.278 & 0.116 & nndsvdar & V matrix & 150 & 0.992 & -0.231 & 0.696 \\
SO & 0.442 & 0.000 & nndsvd & H matrix & 150 & 0.992 & -0.231 & 0.699 \\
SO & 0.500 & 0.000 & random & V matrix & 150 & 0.992 & -0.233 & 0.697 \\
SO & 0.429 & 1.000 & nndsvda & H matrix & 5 & 0.991 & -0.013 & 0.312 \\
SO & 0.183 & 0.000 & nndsvda & both & 150 & 0.991 & -0.227 & 0.698 \\ \bottomrule
\end{tabular}%
}
\caption{Best 5 hyperparameter configurations for NMF on each dataset for IRBO.}
\label{tab:nmf_irbo}
\end{table*}
\section{Computing Infrastructure}
We ran the experiments on a machine equipped with 4 T1390 GPU, CUDA v11.1, 512GB RAM, Intel(R) Xeon(R) CPU E5-2683 v4 @ 2.10GHz.






\end{document}